\documentclass[lettersize,journal]{IEEEtran}

\usepackage{mathtools}
\usepackage{amsmath,amsopn,amssymb}
\usepackage{algorithm}  
\usepackage{algorithmicx}  
\usepackage{algpseudocode}  
\usepackage{amsmath}  

\usepackage{amsthm} 
\usepackage{verbatim}
\usepackage{balance}
\usepackage{amsfonts}
\usepackage{graphicx,xspace,color,soul}
\usepackage{epsfig}
\usepackage{longtable,multirow}
\usepackage{array,float}
\usepackage{algorithmicx}
\usepackage{algorithm}
\usepackage{bm,epstopdf}
\usepackage{tabularx}
\usepackage{breqn}
\usepackage{hyperref}
\usepackage{changepage}

\usepackage{times}
\usepackage{epsfig}
\usepackage{amsfonts}       
\usepackage{booktabs}       
\usepackage{makecell}
\usepackage{algpseudocode}
\usepackage{pifont}

\usepackage{amsmath,amsfonts,bm}

\def\eqref#1{equation~\ref{#1}}

\def\1{\bm{1}}

\DeclareMathAlphabet{\mathsfit}{\encodingdefault}{\sfdefault}{m}{sl}
\SetMathAlphabet{\mathsfit}{bold}{\encodingdefault}{\sfdefault}{bx}{n}

\def\ie{{\it i.e.}}

\ifCLASSOPTIONcompsoc
  \usepackage[nocompress]{cite}
\else
  \usepackage{cite}
\fi

\usepackage{amsmath,amsfonts,amsopn,amssymb,amsthm}
\usepackage{array}
\usepackage[caption=false,font=normalsize,labelfont=sf,textfont=sf]{subfig}
\usepackage{textcomp}
\usepackage{stfloats}
\usepackage{url}
\usepackage{verbatim}
\usepackage{graphicx}
\usepackage{cite}
\usepackage{amsmath,amsfonts,bm}

\def\eqref#1{equation~\ref{#1}}

\def\1{\bm{1}}

\DeclareMathAlphabet{\mathsfit}{\encodingdefault}{\sfdefault}{m}{sl}
\SetMathAlphabet{\mathsfit}{bold}{\encodingdefault}{\sfdefault}{bx}{n}

\usepackage{hyperref}
\usepackage{url}
\usepackage{setspace}

\usepackage{url}

\usepackage{mathtools}
\usepackage{bm}       
\usepackage{algorithm}  
\usepackage{algorithmicx}  
\usepackage{algpseudocode}  
\usepackage[numbers]{natbib}

\usepackage{verbatim}
\usepackage{balance}
\usepackage{graphicx,xspace,color,soul}
\usepackage{longtable,multirow}
\usepackage{array,float}
\usepackage{graphicx}
\usepackage{bm,epstopdf}
\usepackage{tabularx}

\usepackage{times}
\usepackage{epsfig}
\usepackage{booktabs}       
\usepackage{makecell}
\usepackage{pifont}

\makeatletter
\DeclareRobustCommand\onedot{\futurelet\@let@token\@onedot}
\def\@onedot{\ifx\@let@token.\else.\null\fi\xspace}

\def\ie{\emph{i.e}\onedot}

\def\etal{\emph{et al}\onedot}
\makeatother

\def\ourmodel{Multi-scale Latent Point Consistency Models\xspace}
\def\ourmodelshort{MLPCM\xspace}
\usepackage{amsmath,amsfonts}
\usepackage{algorithmicx}
\usepackage{algorithm}
\usepackage{array}
\usepackage[caption=false,font=normalsize,labelfont=sf,textfont=sf]{subfig}
\usepackage{textcomp}
\usepackage{stfloats}
\usepackage{url}
\usepackage{verbatim}
\usepackage{graphicx}
\usepackage{cite}
\begin{document}

\title{Multi-scale Latent Point Consistency Models \\for 3D Shape Generation}

\author{Bi'an Du,~\IEEEmembership{Student Member,~IEEE,}
        Wei Hu,~\IEEEmembership{Senior Member,~IEEE, and}
        Renjie Liao
\thanks{B. Du and W. Hu are with Wangxuan Institute of Computer Technology,
Peking University, No. 128 Zhongguancun North Street, Beijing, China.
E-mail: pkudba@stu.pku.edu.cn, forhuwei@pku.edu.cn}
\thanks{R. Liao is with Department of Electrical and Computer Engineering, University of British Columbia, Vancouver, Canada. E-mail: rjliao@ece.ubc.ca}
\thanks{Corresponding author: Wei Hu.}
}
\pdfgentounicode =1

\markboth{Journal of \LaTeX\ Class Files,~Vol.~14, No.~8, August~2021}%
{Shell \MakeLowercase{\textit{et al.}}: A Sample Article Using IEEEtran.cls for IEEE Journals}


\maketitle

\begin{abstract}
Consistency Models (CMs) have significantly accelerated the sampling process in diffusion models, yielding impressive results in synthesizing high-resolution images. 
To explore and extend these advancements to point-cloud-based 3D shape generation, we propose a novel Multi-scale Latent Point Consistency Model (MLPCM). 
Our MLPCM follows a latent diffusion framework and introduces a hierarchy of latent representations, ranging from point-level to super-point levels, each corresponding to a different spatial resolution. 
We design a multi-scale latent integration module along with 3D spatial attention to effectively denoise the point-level latent representations conditioned on those from multiple super-point levels.
Additionally, we propose a latent consistency model, learned through consistency distillation, that compresses the prior into a one-step generator.
This significantly improves sampling efficiency while preserving the performance of the original teacher model. 
Extensive experiments on standard benchmarks ShapeNet and ShapeNet-Vol demonstrate that MLPCM achieves a 100$\times$ speedup in the generation process, while surpassing state-of-the-art diffusion models in terms of both shape quality and diversity.

\end{abstract}    



\section{Introduction}

Generative 3D shape modeling is fundamental to many 3D vision and graphics applications, enabling digital artists to produce realistic and high-quality assets \cite{luo2021diffusion,zeng2022lion,du2024generative,liu2024one,chen2024sketch2nerf,wang2024hallo3d}. 
For these models to be practically effective, they must provide flexibility for interactive refinement, support the synthesis of diverse shape variations, and generate smooth meshes that seamlessly integrate into standard graphics pipelines.

Driven by rapid progress in text, images, and video generation models, 3D shape generation has likewise advanced progress. Techniques based on variational autoencoders (VAEs)~\cite{tan2018variational,mittal2022autosdf}, generative adversarial networks (GANs) ~\cite{wu2016learning,shu20193d,hao2021gancraft}, and normalizing flow models \citep{yang2019pointflow,klokov2020discrete} have been proposed to generate 3D shapes. Recently, diffusion-based models \citep{mo2023dit} and their latent diffusion variants \citep{zeng2022lion,ren2024xcube} have achieved state-of-the-art results in this domain.


Despite notable progress, substantial obstacles remain when applying diffusion models to 3D shapes that are represented as point clouds. Point sets are unordered, irregular, and often highly non-uniform in density, which complicates both neighborhood construction and feature aggregation \cite{qi2017pointnet,qi2017pointnet++,9925147}. A single radius may fail to capture thin structures while a large radius may blur sharp edges, and k-nearest-neighbor graphs can become unstable under varying sampling density \cite{qi2017pointnet++,wang2019dynamic}. Real data frequently contain noise, outliers, holes, and partial coverage from limited viewpoints \cite{yuan2018pcn,tchapmi2019topnet}, so the denoiser must be robust to missing regions while still reconstructing fine details. In addition, 3D geometry introduces invariances and symmetries that are easy to violate. Models that are not carefully designed with respect to translation, rotation, and scale may overfit to coordinate frames or camera poses \cite{fuchs2020se,satorras2021n}. These factors push the denoising network to learn representations that respect both local geometric primitives such as edges, corners, and small parts, and global object level regularities such as symmetry, topology, and part arrangement. Capturing these patterns jointly is essential for producing high quality point sets that are faithful at a small scale while remaining structurally coherent at a large scale.

A promising strategy is to diffuse in a learned latent space where shapes are summarized into compact codes that suppress nuisance variation and concentrate information about structure \cite{rombach2022high,zeng2022lion}. Learning in latent space reduces computational cost and can improve stability, yet relying on a single level of latent representation has proved insufficient. A single scale cannot simultaneously preserve tiny geometric details and encode global layout, which leads to oversmoothed reconstructions or broken long range dependencies \cite{vahdat2020nvae}. Effective 3D generation therefore requires multiple levels or resolutions together with mechanisms that let information flow across them so that coarse latents guide overall structure while finer latents supply detail where needed \cite{zeng2022lion}. A second challenge concerns practicality. The sampling process in diffusion models typically requires many iterative evaluations of the denoiser, which results in slow generation and high memory consumption \cite{ho2020denoising}. Classifier free guidance and other conditioning signals further increase the computational burden \cite{ho2022classifier}. Without acceleration, these costs hinder real world use in interactive modeling, editing, completion from sparse inputs, or deployment within larger 3D pipelines \cite{song2020denoising,lu2022dpm}. Reducing the number of steps while preserving quality, for example through consistency distillation or higher order solvers in latent space \cite{song2023consistency,lu2022dpm++}, remains critical for making diffusion-based point cloud generators both accurate and efficient.

Here we introduce \ourmodel\ (\ourmodelshort), a generative model for 3D shape synthesis that follows the latent diffusion paradigm while addressing the joint learning of fine local geometry and long range structural coherence within a single framework. The architecture is a hierarchical variational autoencoder that encodes each shape into multiple levels of latent variables, from point level to super-point levels, each aligned with a specific spatial resolution. These latents form an explicit hierarchy in the encoder so that information flows in both top down and bottom up directions. Coarse semantic latents provide guidance for the organization of global structure, while fine latents inject detail that refines and corrects coarse hypotheses. Diffusion operates on point level latents, which concentrates stochasticity where geometric detail resides and keeps higher level latents clean and semantically stable. A multi-scale latent integration module then conditions the denoiser by modulating noisy point latents with super-point latents from several scales. This cross-scale conditioning encourages the emergence of compositional regularities, supports reuse of part-level patterns across categories, and improves data efficiency under sparse or partial observations.

To further strengthen geometric fidelity and controllability, the denoising network integrates a 3D spatial attention mechanism whose bias term derives from pairwise Euclidean distances. This bias respects rigid transformations and encourages attention to concentrate on geometrically proximate regions while preserving the capacity to aggregate information across the entire shape when long range dependencies are beneficial. Training optimizes a hierarchical evidence lower bound for the VAE together with the standard latent diffusion objective. After training, consistency distillation in latent space yields a latent consistency model that preserves the visual quality of the diffusion sampler while reducing the number of refinement steps, which enables rapid sampling for interactive editing and conditional generation. In combination, hierarchical latent organization, multi-scale integration, geometry aware attention, and accelerated inference provide a practical foundation for downstream tasks such as completion from sparse views, structure-preserving editing, controllable synthesis with part level constraints, and conditional reconstruction from partial point clouds, while also improving robustness to noise, distribution shift, and limited supervision.



In summary, our main contributions are as follows:
\begin{itemize}
    \item We propose a novel Multi-scale Latent Point Consistency Model for 3D shape generation, which builds a diffusion model with a hierarchical latent space, leveraging representations ranging from point to super-point levels.
    \item We propose a multi-scale latent integration module along with a 3D spatial attention mechanism to enable the denoiser to focus on salient neighbourhoods while maintaining global consistency, resulting in sharper topology and more uniform point distribution.
    \item We extend consistency distillation to unstructured point‐latent spaces and couple it with multi-scale latent integration, yielding a student that inherits multi-scale semantics and generates 3D shapes in one to four steps without loss of visual or structural fidelity.
    \item Experiments demonstrate that our method outperforms exiting approaches on the widely used ShapeNet benchmark in terms of shape quality and diversity. Moreover, after consistency distillation, our model is 100$\times$ faster than the previous state-of-the-art in sampling.
\end{itemize}
\section{RELATED WORKS}



In this section, we discuss related works on diffusion models, accelerating diffusion models and 3D point cloud generation, respectively. 

\subsection{Diffusion Models}
Diffusion models reach state-of-the-art image synthesis by learning to invert a progressive noising process that corrupts data step by step. 
Under the standard discrete-time setup, the forward diffusion is a fixed Markov chain that progressively corrupts a clean sample into approximately Gaussian noise. The model is then trained to predict the injected noise or the score of the perturbed sample, enabling iterative denoising during sampling \cite{ho2020denoising}. A complementary continuous-time perspective formulates diffusion as an SDE and learns the score of the perturbed data distribution. This view unifies diffusion and score-based generative modeling in a single framework and supports alternative sampling schemes \cite{song2019generative,song2020score}. Maximum likelihood training for continuous-time score models further improves statistical efficiency and connects diffusion training to classic likelihood principles \cite{song2021maximum}.

Scaling laws and architectural choices have made diffusion a robust default for high-fidelity image generation. Hierarchical or two-stage pipelines that combine powerful text encoders with diffusion backbones yield strong semantic control and photorealism \cite{ramesh2022hierarchical,nichol2021glide,shao2024monodiffusion}. Training the generative dynamics in a compact latent space rather than directly in pixel space substantially lowers compute and memory requirements while preserving visual fidelity, and has become a key enabler of high-resolution synthesis \cite{rombach2022high}. Compared with variational autoencoders and generative adversarial networks, diffusion models typically provide more stable optimization, reliable likelihood estimation, and improved mode coverage, mitigating common issues such as posterior collapse in VAEs and mode dropping in GANs \cite{kingma2013auto,sohn2015learning,goodfellow2020generative}. During inference, samples are produced by integrating the learned reverse dynamics from noise toward data, which can be interpreted either as reversing a discrete diffusion chain or integrating an ODE or SDE defined by the learned score \cite{ho2020denoising,song2020score}.

\subsection{Accelerating Diffusion Models}

Despite their success, diffusion models are constrained by slow sampling due to many iterative denoising steps. Training-free acceleration exploits the continuous-time formulation to integrate probability-flow ODE trajectories in fewer steps, including high-order solvers such as DPM-Solver and its improved variants, as well as unified predictor–corrector samplers and adaptive step-size schemes \cite{lu2022dpm,lu2025dpm,zhao2023unipc}. Training-based acceleration modifies objectives or adds light retraining to shorten trajectories, where optimized discretization learns sampling times tailored to a fixed compute budget \cite{watson2021learning}, truncated diffusion shortens the denoising horizon \cite{lyu2022accelerating,zheng2022truncated}, and neural-operator style learners amortize parts of the reverse dynamics \cite{zheng2023fast}. Distillation compresses multi-step teachers into fast students by matching trajectories or terminal outputs, yielding few-step or even single-step generators \cite{salimans2022progressive,meng2023distillation}.

Recent works push one-step and few-step generation further. Consistency Models enforce a mapping that is consistent along time and can be trained either by distillation or from scratch, enabling rapid one-step sampling \cite{song2023consistency}. Follow-up advances improve training stability and data-only training, reduce reliance on perceptual losses, and formalize the theory of consistency training \cite{song2023improved,geng2024consistency}. Stochastic consistency distillation introduces noise-aware consistency to better align with the teacher’s trajectories and improve sample quality at low NFE \cite{liu2025scott}. In parallel, rectified-flow training has been strengthened to compete with distillation in the low-step regime, showing that a single reflow iteration with improved timestep curricula and robust pre-metrics is often sufficient \cite{lee2024improving}. Flow-matching and flow trajectory distillation style approaches provide another route to few-step or one-step sampling by directly parameterizing ODE paths from noise to data \cite{liu2022flow,liu2023instaflow}. There is also growing interest in faster training of diffusion and consistency models themselves, which leverages cross-init consistency phenomena and curriculum-style schedules to cut training cost while retaining generation quality \cite{xu2024towards}. Together, these strategies reduce latency by one to two orders of magnitude, making diffusion practical for interactive applications and deployment under strict compute budgets.

\begin{figure*}
  \centering
  \vspace{-5mm}
  \includegraphics[width=\textwidth]{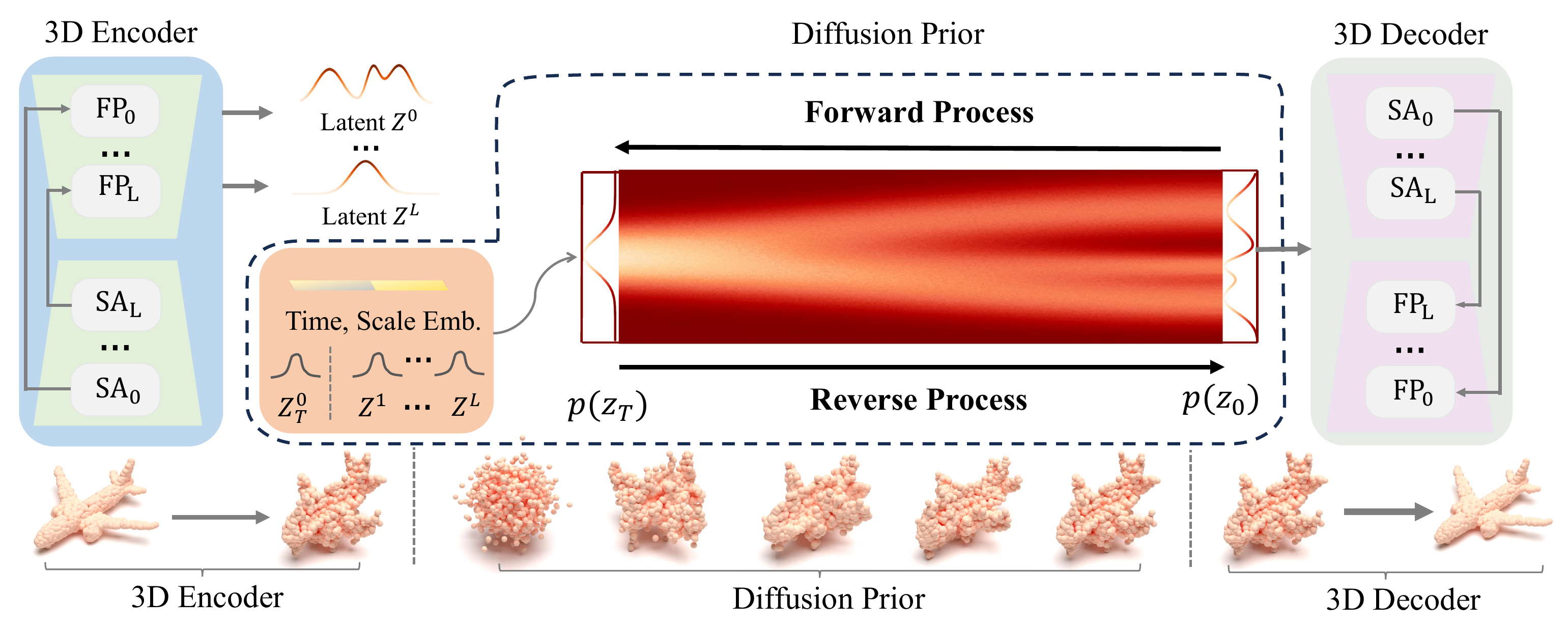}
  \caption{An overview of the proposed multi-scale latent diffusion model. Hierarchical VAEs encode the input shape into latent variables at multiple resolutions, including point-level and super-point-level. Compared to the latent point diffusion model that only denoises the point-level latent space, our proposed multi-scale latent integration module injects higher-level latent variables to guide refinement, thereby jointly modeling local geometry and global structure. The details of the diffusion prior are shown in Figure~\ref{fig:archite_detail}.}
  
  \label{fig:architecture}
\end{figure*}

\subsection{3D Point Cloud Generation}
This setting is commonly framed as 3D point shape and scene generation \cite{yang2019pointflow,zhao2021point,hu2025large,hu2024rangeldm}. Early work by Achlioptas \etal \cite{achlioptas2018learning} formalized the problem with a simple GAN architecture built from stacked MLPs for both the generator and discriminator, and proposed quantitative criteria for assessing the realism of generated point sets. Subsequent studies strengthened the generative backbone by injecting structural inductive biases. Valsesia \etal \cite{valsesia2018learning} introduced graph-convolutional layers to better exploit neighborhood relations, while Liu \etal \cite{liu2018treegan} adopted a tree-structured GCN to encode hierarchical dependencies among points. More recently, Gal \etal \cite{gal2021mrgan} explored a multi-root GAN formulation that can produce point clouds with emergent, unsupervised part-level factorization.

Not all point-cloud generators are GAN-based. A separate stream of work builds on explicit probabilistic formulations \cite{zhang20223dilg,li2023generalized,cheng2023sdfusion}. ShapeGF \cite{cai2020learning}, for instance, defines an (unnormalized) density over 3D space and then uses stochastic gradient ascent to nudge randomly seeded points toward high-density regions, which tend to concentrate around the target surface. Diffusion models offer a different interpretation \cite{10907786,11207156}. DPM \cite{luo2021diffusion} frames generation as running a particle-diffusion process in reverse. They start from noise and apply a shape-conditioned Markov denoising chain, motivated by analogies to thermal diffusion dynamics. There are also flow- and VAE-style approaches. PointFlow \cite{yang2019pointflow} casts a point set as a distribution of distributions and samples via continuous normalizing flows. ChartPointFlow \cite{kimura2021chartpointflow} follows a similar path but introduces label-conditioned charts to better preserve topology. EditVAE \cite{li2022editvae}, in turn, modifies the VAE framework to enable part-aware editing while still supporting unsupervised point-cloud generation.

Several lines of work suggest that plain auto-encoding setups are already quite competitive for learning point-cloud representations and reconstructing shapes sucha as l-GAN \cite{achlioptas2018learning}, ShapeGF \cite{cai2020learning}, and the diffusion-based DPM \cite{luo2021diffusion}. Yet, even with steady progress in 3D generation, the deeper connection between GAN-based modeling and AE-style learning has received little systematic attention.

\section{Multi-scale Latent Point Diffusion Models and Latent Consistency Models}

In this section, we will introduce our hierarchical VAE framework, the multi-scale latent point diffusion prior, and the latent consistency model. 

\subsection{Hierarchical VAE Framework}

    We begin by formally introducing the latent diffusion framework, which is essentially a hierarchical VAE. 
We denote a point cloud as \(X \in \mathbb{R}^{N \times 3} \), consisting of \( N \) points with 3D coordinates. 
We then introduce a hierarchy of latent variables with different spatial scales/resolutions, denoting as $\mathcal{Z} = \{Z^0, Z^1, \dots, Z^L\}$, where $L$ is the number of hierarchy levels.
We use superscript to denote the index of scales.
Here $ Z^l \in \mathbb{R}^{N_l \times C_l}$ where $N_l$ and $C_l$ are the number of points and the number of channels at the $l$-th level latent space respectively.
Note that $C_l=3+D_l$, where $D_l$ represents the extra dimension.
At the $0$-th, we have $N_0 = N$ and $N_l > N_{l+1}$ for any $l = 0, \dots, L-1$.
In other words, the $Z^0$ is latent point representation whereas $Z^l$ are latent super-point (\ie, a subset of points) representations for any $l > 0$. The details of latent layers and points in each level are demonstrated in Section~\ref{Sec:Exp}.
\pdfgentounicode=1

The backbone of our hierarchical VAE consists of an  encoder $q_{\phi}(\mathcal{Z} \vert X)$, a decoder $p_{\psi}(X \vert \mathcal{Z})$, and a prior $p(\mathcal{Z})$, where $\phi$ and $\psi$ are the learnable parameters.
Specifically, our encoder is designed as follows,
\begin{equation}
    q_{\phi}(\mathcal{Z} \vert X) := q_{\phi}(Z^L \vert X) \prod_{l=L-1}^{0} q_{\phi}(Z^l \vert Z^{l+1}, X),
\end{equation}
where $q_{\phi}(Z^L \vert X)$ and $q_{\phi}(Z^l \vert Z^{l+1}, X)$ are assumed to be factorized Gaussian distributions with learnable mean and fixed variance.
Our decoder and prior take the simple form $p_{\psi}(X \vert \mathcal{Z}) := p_{\psi}(X \vert Z^0)$ and $p(\mathcal{Z}) := \prod_{l=0}^{L} p(Z^{l})$ respectively. 
Here $p(Z^{l}) := \mathcal{N}(\bm{0}, I)$.
We model the decoder $p_{\psi}(X \mid Z^0)$ as a fully factorized Laplace likelihood, where the location (mean) parameters are predicted by the network and the scale is fixed to 1. Under this choice, maximizing the reconstruction likelihood is equivalent to minimizing an L1 reconstruction loss.

The overall architecture is illustrated in Figure~\ref{fig:architecture}.
As illustrated, both the encoder and decoder are built from stacked Set Abstraction (SA) and Feature Propagation (FP) modules. Each SA module contains point-voxel convolution (PVC) layers \cite{liu2019point}, followed by a Grouper block, which includes the sampling and grouping layers introduced by \cite{qi2017pointnet++}. 
Each FP module consists of a nearest-neighbor interpolation operation, followed by MLPs and PVC layers.


\textbf{First Stage Training.} We optimize the encoder and decoder pair under a fixed prior by maximizing a modified evidence lower bound (ELBO),
\begin{align}
    \mathcal{L}_{\text{ELBO}}(\phi, \psi)&=\mathbb{E}_{p_{\text{data}}(X),q_{\phi}(\mathcal{Z} \vert X)}[ \text{log} \:p_{\psi}(X \vert \mathcal{Z}) \nonumber  \\
    &- \lambda D_{KL}(q_\phi(\mathcal{Z}|X) \Vert p(\mathcal{Z})) ].
\end{align}

Here $p_{\text{data}}(X)$ represents the unknown data distribution of 3D point clouds.
The hyperparameter \( \lambda \) controls the trade-off between the reconstruction error and the Kullback-Leibler (KL) divergence.

\subsection{Multi-Scale Latent Point Diffusion As Prior}
\pdfgentounicode=1

After learning the encoder and the decoder in the first stage, we fix them and train a denoising diffusion based prior in the latent point space. 
Note that we have a set of latent representations $\mathcal{Z} = \{Z^0, \dots, Z^L\}$ ranging from point to super-point levels output by the hierarchical encoder.
It is natural to select the latent variable with the coarsest scale (lowest spatial resolution), \ie, $Z^L$, to build a diffusion prior since it would enable efficient diffusion in a low-dimensional latent space. 
However, as demonstrated in prior work \citep{zeng2022lion}, point-level latent variables remain crucial for producing high-quality 3D shapes. 
We thus focus on the point-level latent variable $Z^0$ and design mechanisms to fuse multi-scale information. 
Specifically, we introduce a multi-scale latent point diffusion prior as below,
\begin{equation}
    p(\mathcal{Z}) := p_{\theta}(Z^{0} \vert \mathcal{Z}^{\setminus{0}}) \prod_{l=1}^{L} p_{\theta}(Z^{l}),
\end{equation}
where $\mathcal{Z}^{\setminus{0}} = \{ Z^{1}, \dots, Z^{L} \}$.
Here $p_{\theta}(Z^{l})$ is again a factorized standard Normal distribution, whereas $p_{\theta}(Z^{0} \vert \mathcal{Z}^{\setminus{0}})$ is a diffusion model.
This design is crucial for enabling efficient sampling and consistency distillation, as it requires learning only one conditional prior while keeping others fixed.
In contrast, previous work such as LION \cite{zeng2022lion} needs to learn multiple priors corresponding to different levels in the hierarchy, which significantly complicates sampling and distillation.
As in standard diffusion frameworks, our approach comprises a forward noising process and a learned reverse denoising process.

\textbf{Forward Process.} 
Following diffusion models \citep{ho2020denoising}, given initial latent point feature \( Z^0 \sim q_{\phi}(Z^0 \vert X) \) output by the encoder, we gradually add noise as follows, 
\begin{align}
    &q(Z^{0}_{1:T} \vert Z^0, \mathcal{Z}^{\setminus{0}}):=\prod_{t=1}^{T}q(Z^0_t \vert Z^{0}_{t-1}), \\
    &q(Z^{0}_t|Z^{0}_{t-1}):=\mathcal{N}(Z^{0}_t; \sqrt{1-\beta_{t}}Z^{0}_{t-1}, \beta_t I),
\end{align}
where $Z^{0}_{0} := Z^{0}$ and we use subscripts to denote diffusion steps.
\( T \) represents the number of diffusion steps, \( q(Z_t | Z^{0}_{t-1}) \) is a Gaussian transition probability with variance schedule \( \beta_1, \ldots, \beta_T \). We adopt a linear variance schedule in the diffusion process.
The choice of \( \beta_t \) ensures that the chain approximately converges to the stationary distribution, \ie, standard Gaussian distribution \( q(Z_T \vert Z^0, \mathcal{Z}^{\setminus{0}}) \approx \mathcal{N}(Z_T; 0, I) \) after \( T \) steps.

\textbf{Reverse Process.} 
In the reverse process, given the intial noise $Z^0_T \sim \mathcal{N}(Z^0_T; \bm{0}, I)$, we learn to gradually denoise to recover the observed latent point feature $Z^0$ as follows,
\begin{align}
    p_{\theta}(Z^{0}_{0:T} \vert \mathcal{Z}^{\setminus{0}}) &:= p(Z^0_T)\prod_{t=1}^T p_{\theta}(Z^{0}_{t-1} \vert Z^0_t, \mathcal{Z}^{\setminus{0}}), \\ 
    p_{\theta}(Z^{0}_{t-1} \vert Z^0_t, \mathcal{Z}^{\setminus{0}}) &:=\mathcal{N}(Z^{0}_{t-1};\mu_{\theta}(Z^0_t, t, \mathcal{Z}^{\setminus{0}}), \sigma_t^2 I),
\end{align}
where the mean function $\mu_{\theta}(\cdot, \cdot, \cdot)$ of the denoising distribution could be constructed by a neural network with learnable parameters $\theta$.
In particular, following DDPM, we adopt the reparameterization $\mu_{\theta}(Z^0_t, t, \mathcal{Z}^{\setminus{0}}) = (Z^0_t - \beta_t \epsilon_{\theta}(Z^0_t, t, \mathcal{Z}^{\setminus{0}})/ \sqrt{1 - \bar{\alpha}_t}) /\sqrt{\alpha_t}$, where $\alpha_t := 1 - \beta_t$ and $\bar{\alpha}_t := \prod_{s=1}^t \alpha_s$.
The variance $\sigma_t$ is a hyperparameter and set to 1.
Therefore, learning our multi-scale latent point diffusion prior is essentially learning the noise-prediction function $\epsilon_{\theta}(\cdot, \cdot, \cdot)$ parameterized by $\theta$. 

\textbf{Second Stage Training.} 
To train the prior, we ideally aim to minimize the KL divergence between the so-called \emph{aggregated posterior} $q(\mathcal{Z}) = \int p_{\text{data}}(X) q_{\phi}(\mathcal{Z}|X) \mathrm{d}X$ and the prior $p_{\theta}(\mathcal{Z})$.
However, it is again intractable so that we need to resort to the negative ELBO, which can be equivalently written as the following denoising score matching type of objective,
\begin{equation}
    \mathcal{L}_{\text{SM}}(\theta)=\mathbb{E}_{t, X, \mathcal{Z}, \epsilon} \Vert \epsilon - \epsilon_{\theta}(Z^0_t, t, \mathcal{Z}^{\setminus{0}}) \Vert_2^2,
\end{equation}
where the expectation is taken with respect to $t \sim U([T])$, $X \sim p_{\text{data}}(X)$, $\mathcal{Z} \sim q_{\phi}(\mathcal{Z} \vert X)$, $\epsilon \sim \mathcal{N}(\bm{0},I)$.
Here $U([T])$ represents the uniform distribution over $\{1, 2, \dots, T\}$.
Following \cite{zeng2022lion}, we utilize the mixed score parameterization, which linearly combines the input noisy sample and the noise predicted by the network, resulting in a residual correction that links the input and output of the noise function $\epsilon_{\theta}$. 


\begin{figure*}
\begin{center}
    \includegraphics[width=1.0\textwidth]{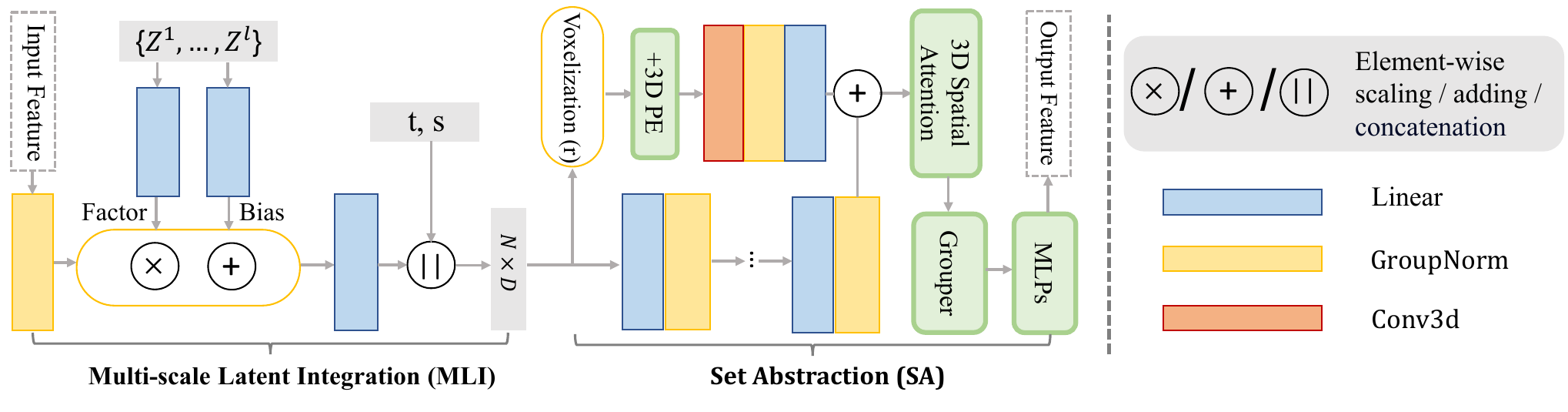}
\end{center}
\vspace{-0.2in}
\caption{An illustration of the network architecture of the latent diffusion prior. The multi-scale latent integration module injects higher-level latents to couple local geometry with global structure, while the set abstraction module downsamples and aggregates across scales. We denote the voxel grid size as $r$, and the hidden dimension as $D$.
}
\label{fig:archite_detail}
\end{figure*}

\subsection{Architecture Design}
\pdfgentounicode=1

We now introduce the specific architecture of our noise prediction function $\epsilon_{\theta}$.
Our backbone network is based on Point-Voxel CNNs (PVCNNs) \citep{liu2019point}.
To integrate multi-scale latent point feature, we propose the multi-scale latent integration (MLI) module before each SA and FP module of PVCNNs. 
We also adopt a 3D spatial attention mechanism in each point-voxel convolution (PVC) module, allowing the model to selectively attend to informative areas, as shown in Figure~\ref{fig:archite_detail}. 

\textbf{Multi-Scale Latent Integration.}
To better capture geometric patterns of 3D shapes, we propose the multi-scale latent integration (MLI) module to fuse latent variables across multiple scales/resolutions.
As shown in Figure~\ref{fig:archite_detail}, at the diffusion step $t$, the $s$-th MLI module takes the feature map $F_{s}$ as input, where $F_{0} = Z^0_t$. 
Specifically, given the high-resolution feature map \(F_s\) and the target feature \(F\), we modulate the normalized target feature \(F\) with the scaling parameters of both the two-dimensional scale and the high-scale features \(F^i\), resulting in an intermediate representation as follows:
\begin{align}
    F_{s+1} & = \left[ \text{MLP} \circ \text{ReLU}( \text{Linear}(\mathcal{Z}^{\setminus{0}}) \odot \text{Norm}(F_{s}) \right. \nonumber \\
    & + \left. \text{Linear}(\mathcal{Z}^{\setminus{0}})) ~\Vert~ \text{Pos}(t) ~\Vert~ \text{Pos}(k) \right],
\end{align}
where Norm is the layer normalization \cite{ba2016layer} and Linear denotes a linear layer.
MLP consists of two linear layers.
$\odot$ denotes element-wise product and $\circ$ means function composition.
$[\cdot \Vert \cdot]$ means concatenation of vectors.
Pos$(\cdot)$ is the positional embedding function.
To enable scale-time awareness in the model, we further modulate the feature with positional embedding of scale and time.
The feature is fed to the following SA and FP modules to predict the noise.

\textbf{3D Spatial Attention Mechanism.}
To help latent point representations better capture 3D spatial information , we design a 3D spatial attention mechanism. 
Our intuition is that closer points in 3D space would have a stronger correlation, thus higher attention values. 
We introduce a pairwise bias term that depends on the 3D distances between points.  
Specifically, in PVC, following \cite{liu2019point}, we first use PointNet++ \cite{qi2017pointnet++} to extract the local features \(F_p\) from the point cloud. 
We then convert the point cloud data into voxels, enabling the application of 3D convolution operations to obtain voxel features \(F_v\). These are then fused to get the combined feature \(F_a=\text{Add}(F_p, \text{MLP}(F_v))\). For \(F_a\), we compute the pairwise spatial distance matrix $B \in \mathbb{R}^{n^{i} \times n^{i}}$ between different point pairs at the current scale, where $n^{i}$ is the number of points at the current scale $i$, and use it as a bias in computing the attention,
\begin{equation}
    F_a = \text{softmax}\left(\frac{QK^T}{\sqrt{d}} + B\right) \cdot V,
\end{equation}
where $Q$, $K$, and $V$ are query, key, and value respectively in the standard attention module.
We add this module to each PVC layer across all scales to make the model's attention conform to the 3D distances between points.
Additionally, following \cite{mo2023dit}, we add 3D positional embeddings to adaptively aggregate feature from voxelized point clouds. 
In each PVC layer, given voxelized input $V\in \mathbb{R}^{v \times v \times v \times c}$, where $c$ is the embedding dimension and $v^3$ denotes the number of voxelized tokens, we add sine-cosine-based 3D positional embeddings to all input tokens.

\subsection{Multi-Scale Latent Point Consistency Models}

Since the diffusion models are notoriously slow in sampling, we explore the consistency models (CMs) \citep{song2023consistency} in the latent diffusion framework to accelerate the generation of 3D shapes.
At the core of CMs, a consistency map is introduced to map any noisy data point on the trajectory of probability-flow ordinary differential equations (PF-ODEs) to the starting point, \ie, clean data. 
There are two ways to train such a consistency map \citep{song2023consistency}, \ie, consistency distillation and consistency training.
The former requires a pretrained teacher model whereas the latter trains the consistency map from scratch.
We explored both options and found consistency distillation works well, whereas consistency training is sensitive to hyperparameters.


Specifically, we first rewrite the reverse process in the continuous-time setting following \cite{song2020score,lu2022dpm}, %

\begin{equation}\label{eq:pf_ode}
\scalebox{0.92}{$
\begin{aligned}
      \frac{\mathrm{d} Z^0_t}{\mathrm{d} t} = \frac{\mathrm{d} \log \alpha_t}{\mathrm{d} t}  Z^0_t + \left( \frac{\mathrm{d} \log \alpha_t}{2\sigma_t \mathrm{d} t} - \frac{\mathrm{d} \log \alpha_t}{\mathrm{d} t} \sigma_t \right) \epsilon_\theta(Z^0_t, t, \mathcal{Z}^{\setminus{0}}), 
\end{aligned}
$}
\end{equation}
where \(\epsilon_\theta(Z^0_t, t, \mathcal{Z}^{\setminus{0}})\) is the noise prediction model and $\alpha_t, \sigma_t$ are determined by the noise schedule as aforementioned. 
Sampling corresponds to integrating the PF-ODE backward from $T$ to $0$. Following \cite{luo2023latent,song2023consistency}, we instead learn a consistency mapping $f_{\hat{\theta}}: (Z^0_t, t, \mathcal{Z}^{\setminus{0}}) \mapsto Z_0^0$ that directly outputs the PF-ODE endpoint $Z^0_0$, enabling consistency distillation. At $t=0$, we express $f_{\hat{\theta}}$ in terms of the noise predictor $\hat{\epsilon}_\theta$:

\begin{equation}
\scalebox{0.92}{$
\begin{aligned}
  f_{\hat{\theta}}(Z^0_t, t, \mathcal{Z}^{\setminus{0}}) = c_{\text{skip}}(t) Z^0_t + c_{\text{out}}(t)\left(\frac{Z^0_t - \sigma_t\hat{\epsilon}_{\hat{\theta}}(Z^0_t, t, \mathcal{Z}^{\setminus{0}})}{\alpha_t}   \right),
\end{aligned}
$}
\end{equation}
where \( c_{\text{skip}}(0) = 1 \), \( c_{\text{out}}(0) = 0 \), and \(\hat{\epsilon}_{\hat{\theta}}(z, c, t)\) is a noise prediction model whose initialized parameters are the same as those of the teacher diffusion model. Note that \( f_{\hat{\theta}} \) can be parameterized in various ways, depending on the teacher model's parameterization of the  diffusion model.

We assume a reliable ODE solver can approximate the integral in Eq.~\ref{eq:pf_ode} over any time interval. The solver is used only during training, not at inference. We then learn to match the PF-ODE solution by minimizing a consistency distillation objective \cite{song2023consistency}:%

\begin{equation}
\scalebox{0.92}{$
\begin{aligned}
  &\mathcal{L}_{\text{CD}}(\hat{\theta}, \hat{\theta}^{-}; \Psi) = \\ 
    &\mathbb{E}_{n,\mathcal{Z}^{\setminus{0}}, Z^0_{t_{n+1}}} \left[ d\left(f_{\hat{\theta}}\left( 
    Z^0_{t_{n+1}}, \mathcal{Z}^{\setminus{0}}, t_{n+1}\right), f_{\hat{\theta}^{-}}\left( 
    \hat{Z}^0_{t_{n}}, \mathcal{Z}^{\setminus{0}}, t_{n}
    \right) \right) \right],
\end{aligned}
$}
\end{equation}
where \( \hat{Z}^0_{t_{n}} = Z^0_{t_{n+1}} + (t_{n} - t_{n+1}) \Psi(Z^0_{t_{n+1}}, t_{n+1}, t_{n}, \mathcal{Z}^{\setminus{0}}) \) is the solution of the \text{PF-ODE} obtained via calling the ODE solver \( \Psi \) from time \( t_{n+1} \) to \( t_n \). Here
$d(\bm{x}, \bm{y}) =  \Vert \bm{x} - \bm{y} \Vert^2_2$ denotes the discrepancy between two latent points, which penalizes large deviations more heavily and thus provides a smooth, scale-aware objective in the latent space.

\section{Experiments}
\label{Sec:Exp}
In this section, we train the multi-scale latent diffusion model on the Shapenet dataset and obtain MLPCM using the latent consistent distillation. 
We begin by describing the dataset and evaluation metrics, and then evaluate MLPCM on the single-class 3D point cloud generation benchmark.
Next, we compare the results of MLPCM with a large number of baselines on multi-class unconditional 3D shape generation in terms of the performance, sampling time, and model parameters. 
Finally, we give a detailed ablation study on the effectiveness of invidual contributions.

\begin{figure*}
\begin{center}
    \includegraphics[width=1.0\textwidth]{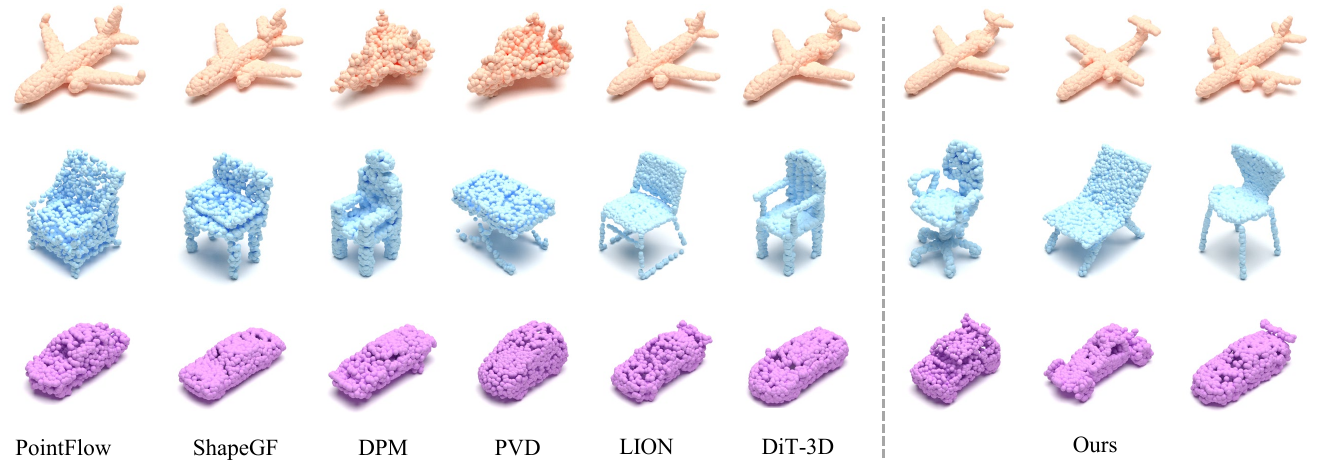}
\end{center}
\vspace{-0.2in}
\caption{We train class-specific ShapeNet models (\textit{airplane}, \textit{car}, \textit{chair}) and generate 2,048-point unconditional samples with PointFlow-style global normalization. Our results better preserve part coherence such as seat–leg and wing–fuselage continuity, while prior methods often show disconnections, fused parts, or symmetry artifacts.}

\label{fig:main_result2}
\end{figure*}

\begin{figure}
\begin{center}
    \includegraphics[width=1.0\columnwidth]{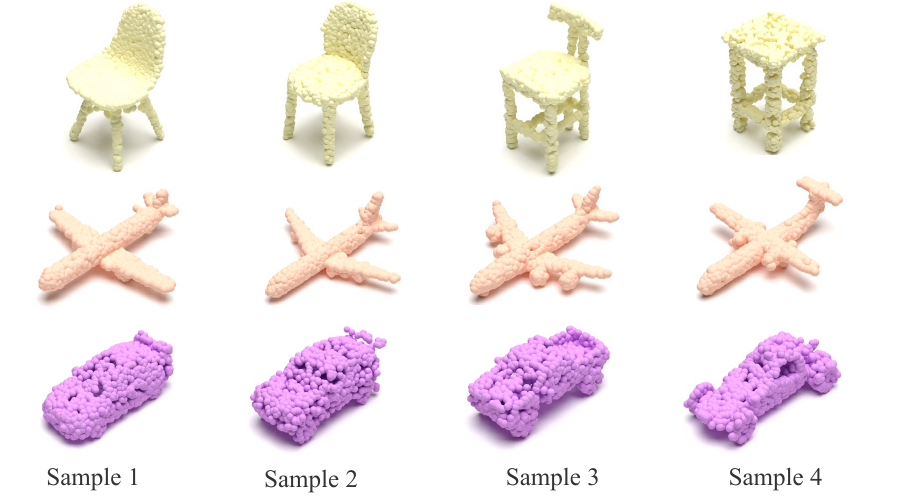}
\end{center}
\vspace{-0.2in}
\caption{Qualitative results show high-quality and diverse 3D assets. Samples maintain realistic structure and overall topology while exploring fine-grained detail changes, yielding diversity with symmetry largely intact.}
\label{fig:main_result}
\end{figure}


\begin{figure}
\begin{center}
    \includegraphics[width=0.9\columnwidth]{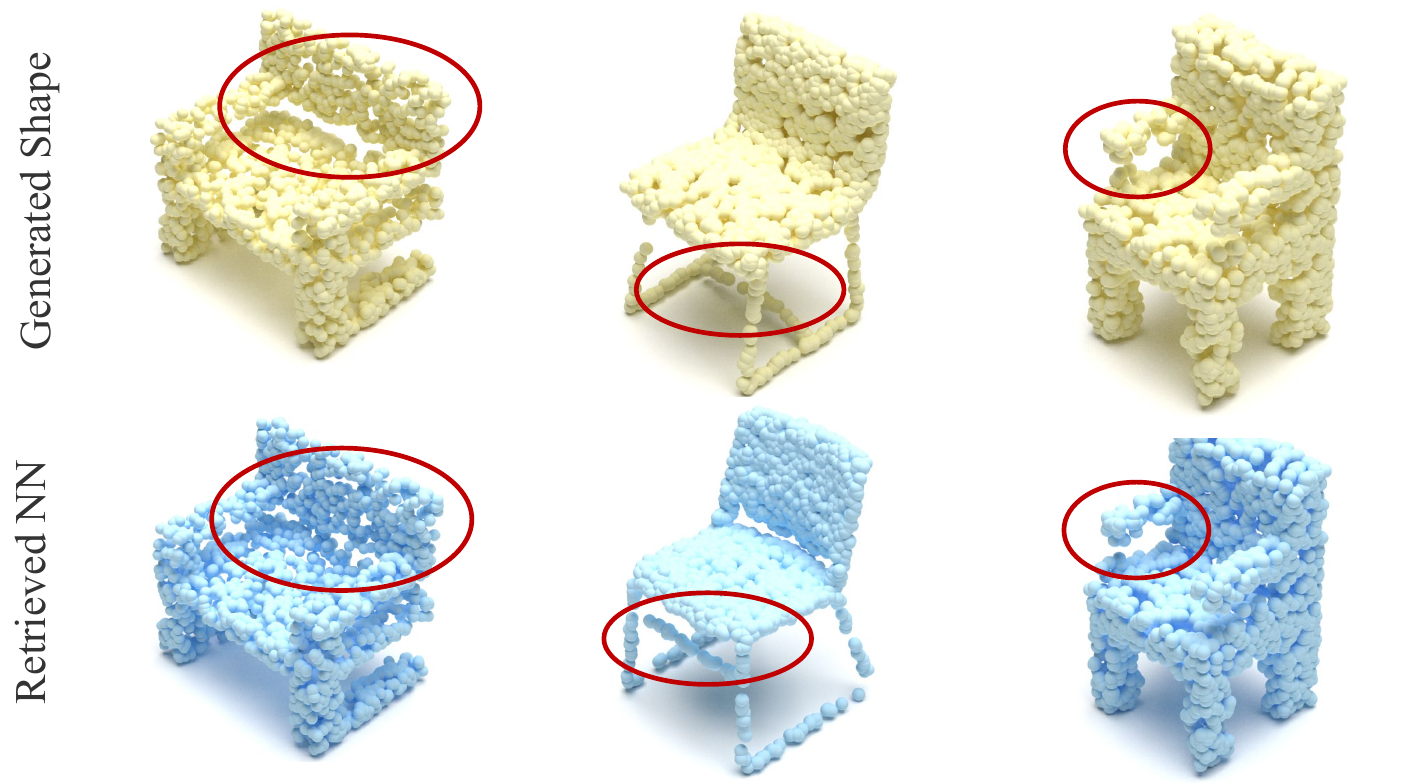}
\end{center}
\vspace{-3mm}
\caption{Comparison between generated shapes and retrieved nearest neighbors (NN) in the training dataset. The generated shapes closely match the ground-truth shapes in overall topology, yet differ in fine-grained details, demonstrating diversity.}
\label{fig:NN1}
\end{figure}

\begin{figure}
    \includegraphics[width=\columnwidth]{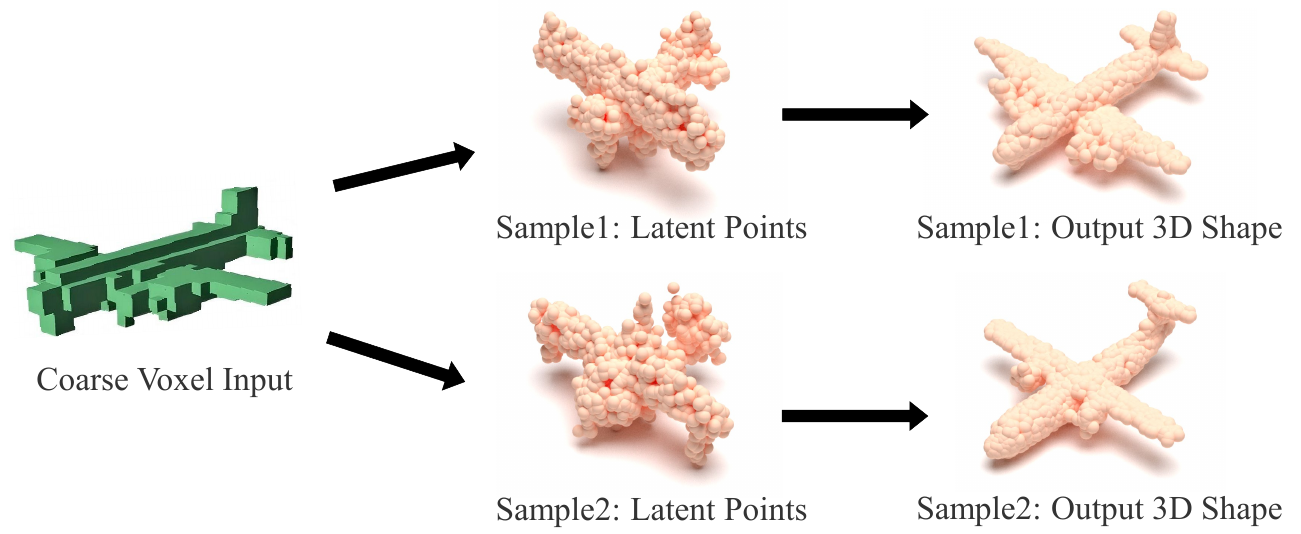}
\caption{Interactive voxel-guided generation. From a coarse occupancy grid, our model samples multiple plausible shapes that respect filled cells while sharpening geometry into smooth surfaces and well-formed parts.}
\label{fig:voxel}
\end{figure}

\subsection{Datasets}

For unconditional 3D point cloud generation, we adopt the ShapeNet benchmark \cite{chang2015shapenet} using the official PointFlow preprocessing pipeline \cite{yang2019pointflow} to keep the data protocol identical to prior work. For a controlled and fair comparison against representative baselines \cite{yang2019pointflow, zhou20213d, zeng2022lion, mo2023dit, luo2021diffusion}, we restrict evaluation to the three standard categories most frequently reported in the literature: \textit{airplane}, \textit{chair}, and \textit{car}.

Each ShapeNet instance is provided as a dense point set with 15,000 points. During optimization, we perform on-the-fly subsampling by uniformly drawing 2,048 points per shape at every iteration, which acts as lightweight data augmentation and matches the common experimental setup. The resulting training split contains 2,832 airplane shapes, 4,612 chair shapes, and 2,458 car shapes, ensuring that all methods are trained and evaluated under the same category scope and point budget.

During our re-implementation of several baselines on the PointFlow ShapeNet splits \cite{yang2019pointflow}, we found that a subset of prior methods \cite{zeng2022lion,li2021sp,zhang2021learning,cai2020learning,hui2020progressive} adopts a \emph{per-shape} normalization protocol rather than a global, dataset-level normalization. Concretely, for each individual shape, they compute axis-wise means to center the point cloud, and then estimate a shape-specific scale using the maximum spatial extent across the three axes. After this centering-and-scaling step, all coordinates are mapped to a bounded cube, typically within $[-1,1]$.

To ensure apples-to-apples comparisons with these methods, we follow the same per-shape normalization procedure in both training and evaluation when reporting results against that group. This detail is non-trivial as different normalization strategies effectively define different learning problems. Global normalization preserves category-level variations in absolute scale and aspect ratios, whereas per-shape normalization removes such information and emphasizes normalized geometry only. As a result, the results across the two protocols are not directly comparable unless the normalization setup is explicitly aligned. We therefore report and interpret numbers only within the same normalization regime to maintain fairness and avoid misleading conclusions.

\subsection{Implementation Details}
\textbf{Training Details}
We implement the teacher model in PyTorch \cite{paszke2019pytorch} and optimize all components with Adam \cite{kingma2014adam}. The VAE is trained for roughly 8,000 epochs, starting from a learning rate of $1\times 10^{-3}$ with a batch size of 128. We set the latent point dimensionality to 4 and use a skip weight of 0.01 to balance the reconstruction pathway during training. For the diffusion prior, we train for about 24,000 epochs with an initial learning rate of $2\times 10^{-4}$. The diffusion process uses 1,000 time steps with a linear schedule. We then train MLPCM for 100K iterations, again with batch size 128, a learning rate of $2\times 10^{-4}$, and an exponential moving average (EMA) decay of $\mu=0.99995$ to stabilize optimization and improve sampling quality. Both the VAE and diffusion prior are trained on 8 NVIDIA A40 GPUs. Finally, our mixed denoising score network is implemented to exactly match the formulations in \cite{zeng2022lion,vahdat2021score}.

\textbf{Evaluation Metrics}
Consistent with prior studies \cite{yang2019pointflow,zhou20213d,zeng2022lion}, we evaluate using 1-NNA under both Chamfer Distance (CD) and Earth Mover’s Distance (EMD). This statistic compares the generated set against the validation distribution, jointly reflecting sample fidelity and diversity \cite{yang2019pointflow}.

\subsection{Single-Class 3D Point Cloud Generation}
\textbf{Implementation}
Our implementation is based on the PyTorch. The input point cloud size is $2048 \times 3$. 
Both our VAE encoder and decoder are built upon PVCNN \cite{liu2019point}, consisting of 4 layers of SA modules and FP modules. 
The voxel grid sizes of PVC at different scales are $32$, $16$, $8$, and $8$, respectively. 
We adopt the Farthest Point Sampling (FPS) algorithm for sampling in the Grouper block, with the sampled center points being $1024$, $256$, $64$, and $16$. 
KNN (K-Nearest Neighbors) is used to aggregate local neighborhood features, with $32$ neighbors for each point in the Grouper. 
We apply a dropout rate of $0.1$ to all dropout layers in the VAE.

\textbf{Results}
We validate our model's results under two settings: normalizing data across the entire dataset and normalizing each shape individually. 
Samples from MLPCM are shown in Figure~\ref{fig:main_result2} and Figure ~\ref{fig:main_result}, and quantitative results are provided in Table~\ref{table:table1} and Table~\ref{table:table2}. For airplanes, our results better preserve thin structures and global symmetry, with crisper wing contours and fewer broken tips. Competing methods often exhibit wing thickening, tail erosion, or rough leading edges, especially in PointFlow and ShapeGF, while DiT-3D improves the silhouette but still shows uneven point density around the tail. For chairs, our method maintains slender legs with consistent spacing and recovers backrest curvature and armrest topology with clear connectivity, while other methods produce floating clusters near the seat or excessively smoothed armrests. Although LION and PVD capture the overall shape, they lose fine perforations and edge sharpness. For cars, our generations show clearly defined wheel arch depressions and fewer surface pits, whereas other methods yield blotchy bodies with collapsed cavities or roof perforations. DPM and ShapeGF in particular display pronounced surface granularity and local self-intersections.

\begin{figure*}
  \centering
  \vspace{-5mm}
  \includegraphics[width=0.9\textwidth]{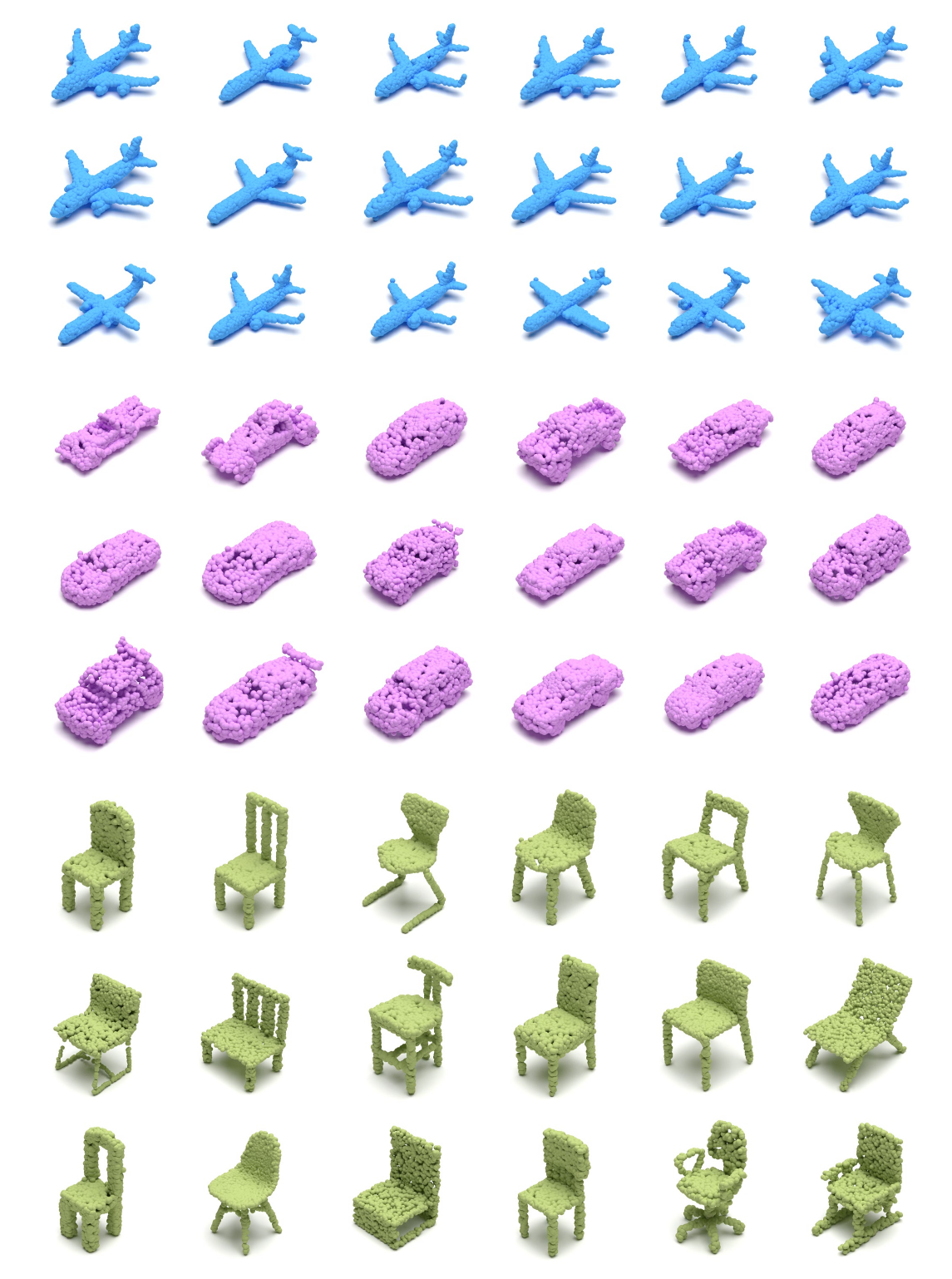}
  \caption{More visualization of Unconditional 3D shape generation via the proposed multi-scale latent consistency model on the airplane, car, and chair categories.}
  
  \label{fig:more_result}
\end{figure*}

More visual results generated by our multi-scale latent consistency model are demonstrated in Figure~\ref{fig:more_result}. Overall, our model achieves sharper thin structures, cleaner topology, and more uniform point distributions, leading to higher structural fidelity than other methods.

Note that TM stands for Teacher Model and LCM stands for Latent Consistency Models.
MLPCM outperforms all baselines and achieves state-of-the-art performance across all categories and dataset versions. 
Compared to key baselines like DPM, PVD, and LION, our samples are diverse and have a better visual quality.

\begin{table*}
\begin{center}
\caption{We evaluate generation on the \textit{airplane}, \textit{chair}, and \textit{car} categories of the PointFlow ShapeNet split \cite{yang2019pointflow} using 1-NNA (lower is better). We apply global normalization to map the train/test sets into $[-1,1]$. TM denotes the Teacher Model, and LCM denotes Latent Consistency Models.}

{
\begin{tabular}{c| cccccc}
\toprule

\multicolumn{1}{c}{} & 
\multicolumn{2}{c}{Airplane} & \multicolumn{2}{c}{Chair} & \multicolumn{2}{c}{Car} 
\\

Method &
CD & EMD & CD & EMD & CD & EMD  \\
\midrule

    r-GAN \cite{achlioptas2018learning}&98.40
         & 96.79 & 83.69 & 99.70  & 94.46 & 99.01   
        \\
    l-GAN (CD)  \cite{achlioptas2018learning}& 87.30
         & 93.95 & 68.58 & 83.84 & 66.49 & 88.78 
        \\
    l-GAN (EMD) \cite{achlioptas2018learning}& 89.49
         & 76.91 & 71.90 & 64.65 & 71.16  & 66.19
        \\
    PointFlow \cite{yang2019pointflow}& 75.68
          & 70.74  & 62.84  & 60.57 & 58.10  & 56.25 \\
    SoftFlow \cite{kim2020softflow}
          &  76.05 & 65.80  & 59.21  & 60.05  &  64.77 & 60.09 
        \\
    SetVAE \cite{kim2021setvae}&76.54
          & 67.65  & 58.84  & 60.57  & 59.94  &  59.94  
        \\
    DPF-Net \cite{klokov2020discrete}
          & 75.18  & 65.55  & 62.00  & 58.53  & 62.35 & 54.48   
        \\
    \midrule
    DPM \cite{luo2021diffusion}& 76.42
          &  86.91 & 60.05  & 74.77  & 68.89  & 79.97 
        \\
    PVD  \cite{zhou20213d}& 73.82
          & 64.81  & 56.26  & 53.32  & 54.55  & 53.83   
        \\
    LION \cite{zeng2022lion}& 67.41
          & 61.23  & 53.70  & 52.34  & 53.41  & 51.14 
        \\
    MeshDiffusion  \cite{liu2023meshdiffusion}& 66.44
          & 76.26  & 53.69 & 57.63  & 81.43  & 87.84 
        \\
    DiT-3D \cite{mo2023dit} & 69.42
          & 65.08  & 55.59  & 54.91  & 53.87  & 53.02
        \\
\midrule
    Ours (TM) &
        \bf65.12  & \bf58.70  & \bf51.48  & \bf50.06 & \bf51.17  & \bf48.92  
        \\
    Ours (LCM)&
        67.22  & 60.32  & 53.63  & 51.74 & 53.82  & 52.75
        \\

\bottomrule

\end{tabular}
}
\label{table:table1}

\end{center}
\end{table*}

\begin{table}
\centering
\caption{We report 1-NNA (lower is better) on the PointFlow ShapeNet benchmark \cite{yang2019pointflow}, using per-shape normalization to map each example into $[-1,1]$.}
\resizebox{\columnwidth}{!}{
\begin{tabular}{c|cccccc}
\toprule
\multicolumn{1}{c}{} &  \multicolumn{2}{c}{Airplane} & \multicolumn{2}{c}{Chair} & \multicolumn{2}{c}{Car}  \\
Method & CD & EMD & CD & EMD & CD & EMD  \\
\midrule
    Tree-GAN \cite{liu2018treegan}&97.53
         & 99.88 & 88.37 & 96.37  & 89.77 & 94.89
        \\
    ShapeGF \cite{cai2020learning}& 81.23
         & 80.86 & 58.01 & 61.25 & 61.79 & 57.24 
        \\
    SP-GAN \cite{li2021sp}& 94.69
         & 93.95 & 72.58 & 83.69 & 87.36  & 85.94
        \\
    PDGN \cite{hui2020progressive}& 94.94
          & 91.73  & 71.83  & 79.00 & 89.35  & 87.22 \\
    GCA \cite{zhang2021learning}
          &  88.15 & 85.93  & 64.27  & 64.50  &  70.45 & 64.20 
        \\
    LION \cite{zeng2022lion}& 76.30 & 67.04 & 56.50 & 53.85  & 59.52  & 49.29         \\
\midrule
    Ours (TM) &
        \bf 73.28 & \bf 63.08  & \bf 56.20 & \bf 53.16 & \bf 58.31 & \bf 47.74         \\
    Ours (LCM)&
        75.56 & 66.85 & 58.58 &  55.32 & 61.28  & 49.91        \\
\bottomrule
\end{tabular}
}

\label{table:table2}
\end{table}

\begin{table}
\begin{center}
\caption{Results of joint train on 13 classes of ShapeNet-vol.}
\resizebox{0.9\columnwidth}{!}
{

{
\begin{tabular}{c|cc}
\toprule


Method &
CD (1-
NNA↓) & EMD (1-
NNA↓)  \\
\midrule

    Tree-GAN \cite{liu2018treegan}&96.80
         & 96.60 
        \\
    PointFlow \cite{yang2019pointflow}& 63.25
         & 66.05
        \\
    ShapeGF \cite{cai2020learning}& 55.65
         & 59.00
        \\
    SetVAE \cite{kim2021setvae}& 79.25
          & 95.25  \\
    PDGN \cite{hui2020progressive}
          & 71.05 & 86.00  
        \\
    DPF-Net \cite{klokov2020discrete}& 67.10
          & 64.75 
        \\
    DPM \cite{luo2021diffusion}& 62.30
          & 86.50
        \\
    PVD \cite{zhou20213d}& 58.65
          & 57.85 
        \\
    LION \cite{zeng2022lion}& 51.85
          & 48.95
        \\
\midrule
    Ours (TM) &
         \textbf{50.17} &  \textbf{47.84} 
        \\
    Ours (LCM)&
         53.85 & 52.45
        \\

\bottomrule

\end{tabular}
}
\label{table:table3}

}
\end{center}
\end{table}

\begin{table*}[]
\centering
\caption{The sampling rates and generation quality of our model with various methods. Our approach achieves high-quality shape generation within 0.5 seconds while maintaining quality. }
{
\begin{tabular}{cc|ccc}
\toprule


Method & steps & time (sec) &
CD (1-
NNA↓) & EMD (1-
NNA↓)  \\
\midrule

    LION (DDPM) \cite{zeng2022lion} & 1000 & 27.09 & 53.41
          &   51.14
        \\
    LION (DDIM) \cite{zeng2022lion} & 1000 & 27.09 & 54.85
          &  53.26
        \\
    LION (DDIM) \cite{zeng2022lion} & 100 & 3.07 & 56.04
          &  54.97
        \\
    LION (DDIM) \cite{zeng2022lion} & 10 & 0.47 & 90.38
          &  95.4
        \\
\midrule
    Ours (TM) & 1000 & 25.16
          &  \textbf{51.17} & \textbf{48.92}
        \\
    Ours (LCM) & 1 & \textbf{0.18}
          &  79.46 & 82.00
        \\
    Ours (LCM) & 4 & 0.31
          &  53.82 & 52.75
        \\

\bottomrule

\end{tabular}
}

  \vspace{-1mm}
  \label{tab:sampling_time}  
\end{table*}
\begin{table}
  \centering
  \caption{Ablation studies on key 3D design components on the car category. PL and SL represent the point-level latents and shape-level latents, respectively, and 3DSA represents the 3D Spatial Attention Mechanism.}
  \resizebox{\columnwidth}{!}{%
  \begin{tabular}{lccc|cc}
    \toprule
    & PL & SL & 3DSA &  CD (1-
NNA$\downarrow$) & EMD (1-
NNA$\downarrow$) 
    \\
    \midrule
    &  & $\checkmark$ & $\checkmark$ & 74.29 & 72.81
    \\
    \midrule
    & $\checkmark$ &  &  $\checkmark$ &  53.40 & 51.96\\
    \midrule
    & $\checkmark$ & $\checkmark$ &  & 53.02  & 50.75 \\
    \midrule
    & $\checkmark$ & $\checkmark$ & $\checkmark$ & \textbf{51.17} & \textbf{48.92}  \\
    \bottomrule
  \end{tabular}
  }
  \vspace{-1mm}
  \label{tab:ablation1}  
\end{table}
\begin{table}
\centering
\caption{We ablate the added latent-point feature dimension $D_l$ on the car category.}
\begin{tabular}{c|cc}
\toprule
Extra Latent Dim &
CD (1-
NNA↓) & EMD (1-
NNA↓)  \\
\midrule
    0 & 54.56
          &   52.79
        \\
\midrule
    1 & \textbf{51.17}
          &  \textbf{48.92}
        \\
\midrule
    3 & 56.88 &  55.27 
        \\
\midrule
    5 & 59.06
          &  51.90
        \\
\bottomrule

\end{tabular}
  \vspace{-1mm}
  \label{tab:ablation_lp}  
\end{table}
\begin{table}  
  \centering
  \caption{Ablation results on the teacher model on the car category evaluate the  backbone choices in the hierarchical VAE and the diffusion prior. For DGCNN, we additionally test different $k$ values in the $k$-NN graph construction.}
  \resizebox{\columnwidth}{!}{%
  \begin{tabular}{lc|cc}
    \toprule
    & Backbone &  CD (1-
NNA$\downarrow$) & EMD (1-
NNA$\downarrow$) \\
    \midrule
    & DGCNN (knn=10) & 67.66 & 57.08\\
    \midrule
    & DGCNN (knn=20) & 68.54 & 57.60 \\
    \midrule
    & PointTransformer & 89.41 &86.52 \\
    \midrule
    & PointTransformerV2 & 80.09 & 79.57 \\
    \midrule
    & PVCNN & \textbf{51.17} &\textbf{48.92}  \\
    \bottomrule
  \end{tabular}
  }
  \label{tab:ablation_backbone}  
\end{table}
\begin{table}  
  \centering
  \caption{Ablation study on 3D positional encoding on the car category for Teacher Model.}
  \resizebox{\columnwidth}{!}{%
  \begin{tabular}{lc|cc}
    \toprule
    & Settings &  CD (1-
NNA$\downarrow$) & EMD (1-
NNA$\downarrow$) \\
    \midrule
    & w/o 3D PE & 51.60 & 49.38 \\
    \midrule
    & with VPE  & 51.17  & \textbf{48.92} \\
    \midrule
    & with cRPE & \textbf{51.10}  & 49.08 \\
    \bottomrule
  \end{tabular}
  }
  \label{tab:ablation_pe}  
\end{table}

\subsection{Multi-Class 3D Shape Generation}

Following \cite{zeng2022lion}, We also perform joint training of the multi-scale latent diffusion model and MLPCM over 13 ShapeNet categories, including airplane, chair, car, lamp, table, sofa, cabinet, bench, phone, watercraft, speaker, display, ship, and rifle. Due to the highly complex and multi-modal data distribution, jointly training a model is challenging.

We validated the model's generation performance on the Shapenet-Vol dataset, where each shape is normalized, meaning the point coordinates are bounded within $[-1,1]$. 
We report the quantitative results in Table~\ref{table:table3}, comparing our method with various strong baselines under the same setting. 
We found that the proposed multi-scale latent diffusion model significantly outperforms all baselines, and MLPCM greatly improves sampling efficiency while maintaining performance.



\subsection{Comparison with Nearest Neighbors and Voxel-guided Shape Synthesis}
It is widely acknowledged that 1-NNA serves as the primary evaluation metric in this field, largely because no superior alternative currently exists. However, it can be misleading, as perfect memorization of the training set would yield an excellent score. Therefore, we adopt a more informative analysis that present the generated shape alongside its retrieved nearest neighbor, highlighting their structural and semantic similarity while also illustrating diversity in finer details. Specifically, Fig.~\ref{fig:NN1} compares our generated shapes (top) with their retrieved Chamfer‐nearest neighbors (bottom). Global topology 
aligns closely, while fine details 
differ, confirming the diversity.

To supplement the conditional generation experiment to observe the generalization diversity of the model, We add voxel guided generation results as shown in Fig.~\ref{fig:voxel}. We begin by voxelizing the training set and fine-tuning our encoder networks to learn representations that can be accurately decoded back into the original shapes. For voxelized shapes handled by our point-cloud networks, we additionally extract surface samples from the voxel grids and use these points as the network inputs.

\subsection{Sampling Time}

The proposed multi-scale latent diffusion model synthesizes shapes using 1000 steps of DDPM, while MLPCM synthesizes them using only 1-4 steps, generating high-quality shapes in under 0.5 seconds. This makes real-time interactive applications feasible. In Table~\ref{tab:sampling_time}, we compare the sampling time and quality of generating a point cloud sample (2048 points) from noise using different DDPM, DDIM \cite{song2020denoising} sampling methods, and MLPCM. When using $\leq 10$ steps, DDIM's performance degrades significantly, whereas MLPCM can produce visually high-quality shapes in just a few steps or even one step.

\subsection{Ablation Study}
 
In this section, we conduct an comprehensive ablation study to disentangle the effects of the main components in our multi-scale latent diffusion model.
Specifically, we examine the impact of introducing multi-scale latent representations to capture both coarse global geometry and fine-grained local details, incorporating a 3D spatial attention mechanism to enable long-range dependency modeling and improved part-to-part coordination in 3D space, and augmenting the latent point representation with additional feature dimensions to increase expressiveness beyond pure coordinates. For each component, we report controlled comparisons under the same training protocol and evaluation metrics, and analyze how these choices affect generation fidelity, structural coherence, and sample diversity.

\textbf{Ablation on the Key Design Components.} We perform the ablation study on the Car category. 
For the configuration without multi-scale latent representations, we consider two scenarios: one using only point-level latent features to capture fine-grained details and another using only shape-level features to focus on global structure. This allows us to assess the impact of each representation scale individually. The model sizes are kept nearly identical across all settings to ensure fair comparison. As shown in Table~\ref{tab:ablation1}, the full model, integrating multi-scale latent representations and the 3D spatial attention mechanism, consistently outperforms the simplified settings across all evaluation metrics. This demonstrates the effectiveness of combining fine-grained local information with global shape context, along with the attention mechanism, in achieving more accurate and coherent reconstructions.

\textbf{Ablation on Extra Dimension for Latent Points.} We ablate the additional latent point dimension $D_l$ on the Car category in Table~\ref{tab:ablation_lp}. 
We experimented with several different additional dimensions for the latent points, ranging from 0 to 5, where $D_l = 1$ provided the best overall performance. 
As the additional dimensions increase, we observe a decrease in the 1-NNA score. 
We thus use $D_l = 1$ for all other experiments.

\textbf{Ablation on different backbones.}
We conduct an ablation study to evaluate different neural network architectures for processing point clouds in the encoder, decoder, and diffusion prior. The results, presented in Table~\ref{tab:ablation_backbone}, focus on the car category using the teacher model, consistent with other ablation studies. We test four popular backbone architectures: Point-Voxel CNN (PVCNN) \cite{liu2019point}, Dynamic Graph CNN (DGCNN) \cite{wang2019dynamic}, PointTransformer \cite{zhao2021point}, and PointTransformerV2 \cite{wu2022point}. Among these, PVCNN demonstrates the best overall performance, making it our architecture of choice.

\textbf{Ablation on 3D Positional Encoding.}
We conduct an ablation study to evaluate different 3D positional encoding methods. In the Voxel Positional Encoding (VPE) approach, we applied frequency-based sine-cosine 3D positional embeddings to the point-voxel branch in the SA module, following the design of DiT-3D \cite{mo2023dit}. For Contextual Relative Positional Encoding (cRPE), inspired by the Stratified Transformer \cite{lai2022stratified}, we enhanced the traditional positional encoding in the attention mechanism by adding a vector representing the positional offset between Query and Key during each attention computation. The experiments were performed on the car category, with results summarized in Table~\ref{tab:ablation_pe}. The findings show that omitting 3D positional encoding leads to significantly poorer generation results. While VPE and cRPE achieve comparable outcomes, we selected VPE as our 3D positional encoding method for its simplicity and effectiveness.

\section{Conclusion}

In this paper, we propose the Multi-Scale Latent Point Consistency Model (MLPCM) to address the task of efficient point-cloud-based 3D shape generation. We first construct a multi-scale latent diffusion model, which includes a point encoder based on multi-scale voxel convolutions, a multi-scale denoising diffusion prior with 3D spatial attention, and a voxel-convolution-based decoder. The diffusion prior model, which integrates multi-scale information, effectively captures both local and global geometric features of 3D objects, while the 3D spatial attention mechanism helps the model better capture spatial information and feature correlations. Moreover, MLPCM leverages consistency distillation to compress the prior into a one-step generator. On the widely used ShapeNet and ShapeNet-Vol datasets, the proposed multi-scale latent diffusion model achieves state-of-the-art performance, and MLPCM achieves a 100$\times$ speedup during sampling while surpassing the state-of-the-art diffusion models in terms of shape quality. In the future, we are interested in extending MLPCM to other 3D representations such as meshes.

{
    \small
    \bibliographystyle{IEEEtran}
    \bibliography{main}
}

\vfill

\end{document}